\documentclass[runningheads]{llncs}

\usepackage[T1]{fontenc}
\usepackage{graphicx}
\usepackage{multirow}
\usepackage{booktabs}
\usepackage{rotating}
\usepackage{makecell}
\usepackage{cite} 
\usepackage{algorithm}%
\usepackage{algorithmic}%
\usepackage[colorlinks,urlcolor=blue,citecolor=blue,linkcolor=blue]{hyperref}
\usepackage{orcidlink}
\usepackage[misc]{ifsym}
\begin{document}
\title{DANAA: Towards transferable attacks with double adversarial neuron attribution}
%
%
\author{Zhibo Jin\inst{1}\orcidlink{0009-0003-0218-1941} \and Zhiyu Zhu\inst{1}\orcidlink{0009-0009-0231-4410} \and Xinyi Wang \inst{2}\orcidlink{0009-0000-5103-011X} Jiayu Zhang\inst{3}\orcidlink{0009-0008-6636-8656} \and Jun Shen \inst{4}\orcidlink{0000-0002-9403-7140} \and Huaming Chen \inst{1}\textsuperscript{\Letter}\orcidlink{0000-0001-5678-472X}}
%
\authorrunning{Z. Jin et al.}
%
\institute{The University of Sydney, Australia \{zjin0915, zzhu2018\}@uni.sydney.edu.au, huaming.chen@sydney.edu.au \and Jiangsu University, China \and Suzhou Yierqi, China \and University of Wollongong, Australia}
\titlerunning{Towards transferable attacks with double adversarial neuron attribution}
%

\maketitle             
\begin{abstract}
While deep neural networks have excellent results in many fields, they are susceptible to interference from attacking samples resulting in erroneous judgments. Feature-level attacks are one of the effective attack types, which targets the learnt features in the hidden layers to improve its transferability across different models. Yet it is observed that the transferability has been largely impacted by the neuron importance estimation results. In this paper, a double adversarial neuron attribution attack method, termed `DANAA', is proposed to obtain more accurate feature importance estimation. In our method, the model outputs are attributed to the middle layer based on an adversarial non-linear path. The goal is to measure the weight of individual neurons and retain the features that are more important towards transferability. We have conducted extensive experiments on the benchmark datasets to demonstrate the state-of-the-art performance of our method. Our code is available at: \href{https://github.com/Davidjinzb/DANAA}{https://github.com/Davidjinzb/DANAA}.

\end{abstract}
\section{Introduction}
Deep neural networks (DNNs) have been used in a wide range of applications in different fields, such as face recognition~\cite{deng2019arcface}, voice recognition~\cite{aizat2020identification} and sentiment analysis~\cite{wadawadagi2020sentiment}. DNNs can also achieve state-of-the-art performance in tasks such as security verification in unconstrained environments where very low false positive rate metrics are required~\cite{deng2019arcface}. However, deep learning models are shown to be vulnerable to interference from adversarial samples. Attackers can manipulate the model outcome by deliberately adding the perturbations to the original samples to attack the models~\cite{szegedy2013intriguing}. 

In general, the current approaches to attack models can be categorised into two types: white-box attack~\cite{goodfellow2014explaining} and black-box attack~\cite{papernot2017practical}. For white-box attacks, the attacker knows the relevant parameters of the target model and can formulate the most suitable attack method. For black-box attacks, on the other hand, the attacker does not have access to the model parameters. In terms of the characteristics of the white-box and black-box attack methods, the black-box attack provides the adversarial performance of the attacking samples, which is useful for improving the robustness of deep learning models in real-world scenarios. Specifically, the black-box attack methods have three types, including query-based method~\cite{ilyas2018black}, transfer-based method~\cite{dong2019evading} and hybrid method~\cite{fu2022boosting}.

The objective of the query-based method is to interrogate the model to extract pertinent input or output information, and subsequently utilize this limited information to iteratively generate optimal adversarial samples. However, such method is subject to restrictions imposed by access permissions and often require multiple queries to obtain excellent adversarial samples. The transfer-based method aims to train and generate adversarial samples on a known-information local surrogate model, which are then transferred and tested on the target black-box model for the attack success rate. Compared to query-based methods, transfer-based methods do not require additional access to the model and can bypass certain adversarial defense mechanisms aimed at queries. The hybrid method combines the principles of query and transfer approaches. Although it can achieve sufficiently high attack success rate, it also implies that it is susceptible to adversarial defense mechanisms targeting both queries and transfers. Therefore, in this paper, we focus on transfer-based method.

As a common approach of transfer-based attack, feature-level attack attempts to maximise the internal feature loss by attacking intermediate layers' features to improve the transferability of the attack~\cite{wang2021feature}. The aim is to increase the weight of negative features in the middle layer of the model while decreasing the weight of positive features. More negative features will be retained to assist the diversion of the model's predictions. However, it is still challenging to harmoniously differentiate the middle-level features via feature-level attack method, which is also prone to its local optimum~\cite{wang2021feature}. Moreover, it is well-known that the effectiveness of transfer-based black-box attacks is influenced by the overfitting on surrogate models and specific adversarial defenses. To address these challenges, we propose to utilise the information of neuron importance estimation for the middle layer to identify the adversarial features more accurately. In addition, we also evaluate the transferability of our proposed method on adversarially trained models, which will be specifically discussed in Section~\ref{sec:Experiments}. The results demonstrate that our method achieves favorable attack success rates even on target models protected by adversarial defenses.
  
To obtain adversarial samples with higher transferability, this paper presents a double adversarial neuron attribution attack (DANAA). DANAA method attributes the model outputs to the middle layer neurons, thus measuring individual neuron weights and retaining features that are more important towards transferability. We use adversarial non-linear path selection to enrich the attacking points, which improves the attribution results. Extensive experiments on the benchmarking datasets following the literature methods have been conducted. The results show that, DANAA can achieve the best performance for the adversarial attacks. We anticipate this work will contribute to the attribution-based neuron importance estimation and provides a novel approach for transfer-based black-box attack. Our contributions are summarised as follows:
\begin{itemize}
    \item We propose DANAA, an innovative method of non-linear gradient update paths to achieve a more accurate neuron importance estimation, for a more in-depth study of the route to attribution method. 
    \item We present both theoretical and empirical investigation details for the attribution algorithm in DANAA, which is a core part of the method, in Section~\ref{sec:method}.
    \item A comprehensive statistical analysis is performed based on our benchmarking experiments on different datasets and adversarial attacks. The results in Section~\ref{sec:Experiments} demonstrates the state-of-the-art performance of DANAA method.
\end{itemize}

\section{Related Work}
In this section, we review the literature on white-box attacks, query-based black-box attacks, transfer-based black-box attacks, and hybrid black-box attacks.


\subsection{Common white-box attacks}
Previous work has demonstrated that neural networks are highly susceptible to misclassification by pre-addition of perturbed test samples. Such processed samples are called adversarial samples. The emergence of adversarial samples has led to the development of a range of adversarial defences to ensure the model performance~\cite{szegedy2013intriguing}~\cite{kurakin2016adversarial}~\cite{tramer2017ensemble}.

Currently, adversarial attacks can be divided into white-box attacks and black-box attacks depending on the level of available information for the model being attacked. There are various approaches for white-box attacks, such as gradient-based and GAN-based. Gradient-based white-box attacks include FGSM ~\cite{goodfellow2014explaining}, I-FGSM~\cite{kurakin2016adversarial}, PGD~\cite{madry2017towards} and $C\&W$~\cite{carlini2017towards}. Some recent GAN-based white-box attack methods are AdvGAN~\cite{xiao2018generating}, GMI~\cite{zhang2020secret}, KED-MI~\cite{chen2021knowledge} and $Plug\&Play$~\cite{struppek2022plug}. While white-box attacks are effective in measuring the robustness of a model under attack, in real-world scenario, the parameters of the model are often not accessible, leading to the development of black-box attacks.

\subsection{Query-based Black-box Attacks}
Query-based attacks are a branch of black-box attacks aiming to train an effective adversarial sample by performing a small-scale attack on the target model to query the model parameters, such as the model labels and confidence levels. These parameters can be used as part of the dataset to assist in training the migration algorithm to verify the migration of the black-box model. Ilyas et al.~\cite{ilyas2018black} were the first to propose a query-based black-box attack approach. Following, they proposed combining prior and gradient estimation of historical queries and data structures based on Bandit Optimization, which greatly reduces the number of queries~\cite{ilyas2018prior}. Li et al.~\cite{li2020qeba} proposed a query-efficient boundary-based black box attack method (QEBA). It proved that the gradient estimation of the boundary-based attack over the entire gradient space is invalid in terms of the number of queries. Andriushchenko et al.~\cite{andriushchenko2020square} proposed the square search attack method, which selects local square blocks at random locations in the image to search and update the direction of the attack.
\subsection{Transfer-based Black-box Attacks}
 The transferability of adversarial attacks refers to the applicability of the adversarial samples generated by the local model to the target model for attack. The attacker firstly uses the parameters obtained from the attack on the local model to train the adversarial samples, then uses these samples to perform a black-box attack on the target model to verify the success rate. 
 
 There are three main categories of transfer-based black-box attacks, namely gradient calculation methods, input transformation methods and feature-level attack methods. Gradient calculation methods such as MIM~\cite{dong2018boosting}, VMI-FGSM~\cite{wang2021enhancing} and SVRE~\cite{xiong2022stochastic} improve transferability by designing new gradient updates. Input transformation methods such as DIM~\cite{xie2019improving}, PIM~\cite{gao2020patch} and SSA~\cite{long2022frequency} boost the transferability by using input transformations to simulate the ensemble process of the model, while feature-level attacks focus on the middle-layer features. 
 
 Some state-of-the-art feature-level attack methods include NRDM, FDA, FIA and NAA, etc. NRDM~\cite{naseer2018task} attempts to maximise the degree of distortion between neurons, but it does not take into account the role of positive and negative features in the attack. FDA~\cite{ganeshan2019fda} averages the neuronal activation values to obtain an estimate of the importance of a neuron. However, this method does not distinguish the degree of each neuron's importance and the discrimination between positive and negative features is still too low. FIA~\cite{wang2021feature} multiplies the activation values of neurons and back-propagation gradients for estimation, but its effect on the original input is affected by over-fitting and the results are not accurate. NAA~\cite{zhang2022improving} effectively improves the transferability of the model and reduces computational complexity by attributing the model's output to an intermediate layer to obtain a more accurate importance estimation. However, its attribution method focuses more on the gradient iteration process considering linear path, and there is still room for improvement in the non-linear path condition.
 \subsection{Hybrid black-box attacks}
Hybrid method is a combination of query-based method and transfer-based method. It not only considers the priori nature of the transfer but also utilizes the gradient information obtained from the query, which resolves the challenges of high access cost for the query attack and low accuracy for transfer attack. 

Dong et al.~\cite{cheng2019improving} proposed a hybrid method named P-RGF, which used the gradient of surrogate model as prior knowledge to guide the query direction of RGF and obtained the same success rate as RGF with fewer queries. Fu et al.~\cite{fu2022boosting} train Meta Adversarial Perturbation (MAP) on an surrogate model and perform black-box attacks by estimating the gradient of the model, which has good transferability and generalizability. Ma et al.~\cite{ma2021simulating} introduced Meta Simulator to black-box attacks based on the idea of meta-learning. By combining query and transfer based attacks, the researchers not only significantly reduce the number of queries, but also reduce the complexity of queries by transferring the adversarial samples trained on the surrogate model to the target model. 

While there are different types of black-box attack methods, transfer-based attacks is considered as the most convenient method which doesn't require additional information queries for the model. However, it poses the challenge of a good transferability for the adversarial samples. Therefore, in this work, we target the transfer-based attack methods. Especially, we introduce the attribution method for the middle-layer feature estimation, which shows a promising performance with our experiments. 

\section{Method}\label{sec:method}
\subsection{Preliminaries}
When an adversarial attack to the target model can be successfully launched given an adversarial samples trained with a local DNN model, we consider there is a strong transferability relationship between these two models. Formally, with a deep learning network $N:R^{n}\to R^{c}$ and original image sample $x^{0}\in R^{n}$, whose true label is $t$, if the imperceptible perturbation $\sum_{k=0}^{t-1}\bigtriangleup x^{k}$ is applied on the original sample $x^{0}$, we may mislead the network $N$ with the manipulated input $x^{t}=x^{0}+\sum_{k=0}^{t-1}\bigtriangleup x^{k}$ to the label of $m$, which can also be denoted as $x^{adv}$. Assuming the output of the sample $x$ as $N(x)$, the optimization goal will be:
 \begin{equation}
     \left \| x^{t}-x^{0} \right \| _{n}<\epsilon  \quad subject \ to\quad N(x^{t})\ne N(x^{0})
 \end{equation}
Where $\left \| \cdot \right \|_{n} $ represents the n-norm distance. Considering the activation values in the middle layers of network $N$, we denote the  activation value of $y$-th layer as $y$ and the activation value of $j$-th neuron as $y_{j}$. 
\subsection{Non-linear Path-based Attribution}
\begin{figure}[h] 
\centering 
\includegraphics[width=0.7\textwidth]{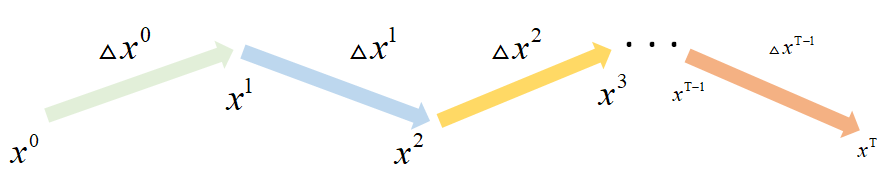}
\caption{Non-linear gradient update path diagram} 
\label{path} 
\end{figure}

Inspired by~\cite{sundararajan2017axiomatic} and~\cite{zhang2022improving}, we define the attribution results of input image $x^{t}$(with $n \times n$ pixels) as 
\begin{equation}
    A:=\sum_{i=1}^{n^{2}} \int\bigtriangleup x_{i}^{t} \frac{\partial N(x^{t} )}{\partial x_{i}^{t} }\mathrm{d}t 
\label{eq2}
\end{equation}
As shown in Fig.~\ref{path}, different from the NAA algorithm~\cite{zhang2022improving}, our paper proposes a new attribution idea that uses a non-linear gradient update path instead of the original linear path, which allows the model to find the optimal path against the attack itself. In Eq.~\ref{eq2}, the gradient of $N$ iterates along the non-linear path $x^{t}=x^{0}+\sum_{k=0}^{t-1}\bigtriangleup x^{k}$, in which $\frac{\partial N}{\partial x_{i}^{t}}(\cdot )$ is the partial derivative of $N$ to the $i$-th pixel. For each iteration, $\bigtriangleup x^{t}=lr \cdot sign(\frac{\partial N(x^{t})}{\partial x_{i}^{t}})+N(0,\sigma)$. We further apply the learning rate and Gaussian noise to update the perturbation. 

Afterwards, we can approximate $A$ as $N(x)$ depending on basic advanced mathematics and extend the attribution results to each layer. The formula of attribution can then be expressed as: 
\begin{equation}
    A_{y_{j}}:=\sum_{i=1}^{n^{2}} \int\bigtriangleup x_{i}^{t} \frac{\partial N(x^{t})}{\partial y_{j}({x^{t}})}\frac{\partial y_{j}(x^{t})}{\partial x_{i}^{t}}\mathrm{d}t
\label{eq3}
\end{equation}
Where $A_{y_{j}}$ represents the attribution of $j$-th neuron in the layer $y$, $\sum A_{y_{j}}=A$. We provide the relevant proof of our non-linear path-based attribution in following section.
\subsection{Proof of Non-linear Path-based Attribution}
Since we now have $A_{y_{j}}$ as Eq.~\ref{eq3}, assuming that the neurons on the middle layer of the deep neural network are independent from each other, $A_{y_{j}}$ can be expressed as
\begin{equation}
    A_{y_{j}}:=\int\frac{\partial N(x^{t})}{\partial y_{j}(x^{t})}\sum_{i=1}^{n^{2}} \bigtriangleup x_{i}^{t} \frac{\partial y_{j}(x^{t})}{\partial x_{i}^{t}}\mathrm{d}t
\label{eq4}
\end{equation}
Where $\frac{\partial N(x^{t})}{\partial y_{j}(x^{t})}$ is the gradient of $N(x^{t})$ to the $j$-th neuron, $\sum_{i=1}^{n^{2}} \bigtriangleup x_{i}^{t} \frac{\partial y_{j}(x^{t})}{\partial x_{i}^{t}}$ is the sum of the gradient of $y_{j}$ to each pixel on $x^{t}(x^{t}\in R^{n})$. Since the two gradient sequences are zero covariance, we then convert Eq.~\ref{eq4} into:
\begin{equation}
    A_{y_{j}}:=\int\frac{\partial N(x^{t})}{\partial y_{j}(x^{t})}\mathrm{d}t\cdot\int \sum_{i=1}^{n^{2}} \bigtriangleup x_{i}^{t} \frac{\partial y_{j}(x^{t})}{\partial x_{i}^{t}} \mathrm{d}t
\label{eq5}
\end{equation}
Combining the principles of calculus, we can prove that 
\begin{equation}
    \int \sum_{i=1}^{n^{2}} \bigtriangleup x_{i}^{t} \frac{\partial y_{j}(x^{t})}{\partial x_{i}^{t}} \mathrm{d}t=y_{j}^{t}-y_{j}^{0}    
\end{equation}
then we denote $y_{j}^{t}-y_{j}^{0}$ as $\bigtriangleup y_{j}^{t}$, Eq.~\ref{eq5} can be converted into
\begin{equation}
    A_{y_{j}}:=\bigtriangleup y_{j}^{t}\int\frac{\partial N(x^{t})}{\partial y_{j}(x^{t})}\mathrm{d}t
\end{equation}
Denoting $\int\frac{\partial N(x^{t})}{\partial y_{j}(x^{t})}\mathrm{d}t$ as $\gamma(y_{j})$, which means the gradient of network $N$ along our non-linear path with attention to the $j$-th neuron. Afterwards, we can get $A_{y_{j}}=\bigtriangleup y_{j}^{t}\cdot \gamma(y_{j})$. Since the neuron $y_{j}$ is on the middle layer $y$, finally the attribution result of the layer $y$ can be expressed as
\begin{equation}
    A_{y}=\sum _{y_{j}\in y}A_{y_{j}}=\sum _{y_{j}\in y}\bigtriangleup y_{j}^{t}\cdot \gamma(y_{j})=\bigtriangleup y^{t}\cdot \gamma(y)
\end{equation}

\begin{algorithm}[htbp]
    \renewcommand{\algorithmicrequire}{\textbf{Require:}} 
    \caption{Double Adversarial Neuron Attribution Attack}
    \label{alg1}
    \begin{algorithmic}[1] 
        \REQUIRE Deep network N, target layer y
        \REQUIRE Manipulated input $x^{t}$ with label m
        \REQUIRE Perturbation budget $\epsilon$ and iteration number T
        \REQUIRE Original input $x^{0}$ and integrated step $\tau$
        \STATE $\alpha=\frac{\epsilon}{T}$, $\gamma(y_{j})=0$, $g_{0}=0$, $\mu=1$, $x^{adv}_{0}=x^{t}$
        
        \FOR {$t=0 \gets \tau$}

            \STATE $x^{t+1}=clip_{x}^{\epsilon}\{x^{t}+lr \cdot sign(\frac{\partial N(x^{t})}{\partial x^{t}})+N(0,\sigma)\}$            
            \STATE  $\gamma(y_{j})=\gamma(y_{j})+\bigtriangledown_{y(x^{t})}N(x^{t})$  
        \ENDFOR
        \FOR {$ s=0 \gets T-1$}
            \STATE $A_{y}=\bigtriangleup y^{t}\cdot \gamma(y)$
            \STATE $g_{s+1}=\mu \cdot g_{s}+\frac{\bigtriangledown_{x^{t}}A_{y}}{\left \| \bigtriangledown_{x^{t}}A_{y} \right \|_{1} }$
            \STATE $x^{adv}_{s+1}=Clip_{x^{t}}^{\epsilon}\left \{ x^{adv}_{s+1}+\alpha \cdot sign(g_{s+1}) \right \} $
        \ENDFOR
    \end{algorithmic} 
\end{algorithm}
Alg.~\ref{alg1} shows the specific pseudocode structure of our DANAA algorithm with Non-linear Path-based attribution.
\section{Experiments}\label{sec:Experiments}

Extensive experiments have been conducted to demonstrate the efficiency of our method. Following sections cover the topic of leveraged datasets, benchmarking models and incorporated metrics. We also provide the experimental settings. We performed five rounds of benchmarking experiments to compare our algorithm with other methods, demonstrating the superiority of our approach to the baselines in terms of transferability for adversarial attacks. Moreover, we conducted the ablation study to investigate our approach, focusing on the impact of various learning rates and noise deviation on attack transferability.

\subsection{Dataset}
Following other literature methods, the widely-used datasets from NAA work~\cite{zhang2022improving} are considered in this paper. The datasets consist 1000 images of different categories randomly selected from the ILSVRC 2012 validation set~\cite{russakovsky2015imagenet}, which we called a multiple random sampling(MRS) dataset.

\subsection{Model}
We include four widely-used models for image classification tasks, namely Inception-v3 (Inc-v3)~\cite{szegedy2016rethinking}, Inception-v4 (Inc-v4)~\cite{szegedy2017inception}, Inception-ResNet-v2 (IncRes-v2)~\cite{szegedy2017inception}, and ResNet-v2-152 (Res152-v2)~\cite{he2016deep}, as source models for assessing the attacking performance of our algorithm. We start with four pretrained models without adversarial learning, which include Inc-v3, Inc-v4, IncRes-v2, and Res152-v2. Later on, we construct more robust models for a in-depth comparison, such as including adversarial training for the pretrained models. This results in two adversarial trained models, including Inception-v3(Inc-v3-adv) and Inception-Resnet-v2 (IncRes-v2-adv)~\cite{kurakin2016adversarial}. The remaining three models are based on the ensemble models: the ensemble of three adversarial trained Inception-v3(Inc-v3-adv-3), the ensemble of four adversarial trained Inception-v3 (Inc-v3-adv-4), and the ensemble of three adversarial trained Inception-Resnet-v2 (IncRes-v2-adv-3), following the work from \cite{tramer2017ensemble}. In \cite{tramer2017ensemble}, the models are combined by training the sub-models of the corresponding model independently and finally weighting the results of each sub-model to increase the accuracy and robustness of the model.

\subsection{Evaluation Metrics}
The attack success rate is selected as the metric to evaluate the performance. It measures the proportion of the dataset where our method produces incorrect label predictions after attacking. Hence, a higher success rate indicates improved performance of the attack method.

\subsection{Baseline methods}
For comparison in our experiment, we selected five state-of-the-art attack methods as the baseline, including MIM~\cite{dong2018boosting}, NRDM~\cite{naseer2018task}, FDA~\cite{ganeshan2019fda}, FIA~\cite{wang2021feature}, and NAA~\cite{zhang2022improving}. Furthermore, to test the effect of each model after combining input transformation methods and to verify the superiority of our algorithm, we apply both DIM and PIM to the attack methods. The implementation details can be found in the open source repository. Consequently, we extend the model comparison set with MIM-PD, NRDM-PD, FDA-PD, FIA-PD, NAA-PD and DANAA-PD, respectively.

\subsection{Parameter Setting}
In the experiment, we set the parameters as following: the learning rate (lr) is 0.0025; the noise deviation is 0.25; and the maximum perturbation rate is 16, which is derived from the number of iterations (15) and the step size (1.07). The batch size is 10, and the momentum of the optimization process is 1. Since we introduced the DIM and PIM algorithms to verify the superiority of our model when combining input transformation methods, we set the transformation probability of DIM to 0.7, and the amplification factor and kernel size of PIM  are 2.5 and 3, respectively. For the target layer of the attack, we choose the same layer as in NAA. Specifically, we attack InceptionV3/InceptionV3/Mixed\_5b/concat layer for Inc-v3; InceptionV4/InceptionV4/Mixed\_5e/concat layer for Inc-v4; InceptionResnetV2/Ince-ptionResnetV2/Conv2d\_4a\_3x3/Relu layer for IncRes-v2; the ResNet-v2-152/blo-ck2/unit\_8/bottleneck\_v2/add layer of Res152-v2~\cite{zhang2022improving}.

\subsection{Result}
All the experiments are carried out with the hardware of RTX 2080Ti card. A detailed replication package can be found in the open source repository at \href{https://github.com/Davidjinzb/DANAA}{https://github.com/Davidjinzb/DANAA}. We subsequently compile the results of all the attack methods without and with the input transformation methods (ending with PD) in Table.~\ref{tab1}. 

In Table.~\ref{tab1}, we can see that, DANAA has retained a strong and robust performance across all the models, in comparison with other attack methods. Especially, DANAA demonstrated notable improvements on five models that are adversarial trained. We can observe a largest improvement of the attacking performance is between our method and NAA method~\cite{zhang2022improving}, which is the generally second best attacking method in the comparison experiments. The ratio of improvement is 9.0\%. Across all local models, our approach demonstrated an overall average improvement of 7.1\% as compared to NAA on the adversarial trained models. By introducing the PD concept, our method achieves a maximum improvement of 9.8\% over NAA-PD and an overall average improvement of 7.3\% on the adversarial trained models.

\begin{table}[htbp]
  \centering
  \caption{Attack success rate of multiple methods on different models}\label{tab1}
  \begin{tabular}{l|c|ccccccccc}
    \toprule
    \hline
    Model & Attack method & Inc-v3 & Inc-v4 & \makecell{IncRes\\-v2} & \makecell{Res152\\-v2} & \makecell{Inc-v3\\-adv} & \makecell{IncRes\\-v2-adv} & \makecell{Inc-v3\\-adv-3} & \makecell{Inc-v3\\-adv-4} & \makecell{IncRes-v2\\-adv-3} \\
    \hline
    \multirow{6}[2]{*}{Inc-v3} & MIM & \textbf{100} & 41.9 & 39.7 & 32.8 & 22.1 & 18.4 & 14.9 & 15.7 & 8.2 \\
          & NRDM & 90.4 & 61.4 & 52.5 & 49.9 & 26.1 & 19.2 & 9.5  & 12.9 & 4.7 \\
          & FDA  & 81.7 & 42.9 & 37.1 & 35.1 & 19.4 & 12.6 & 9.3  & 12.2 & 5.0 \\
          & FIA  & 96.5 & 79.1 & 77.8 & 71.8 & 54.8 & 53.9 & 43.1 & 44.2 & 23.2 \\
          & NAA  & 97.0   & 83.0   & 80.6 & 74.7 & 56.2 & 59.4 & 49.5 & 50.4 & 31.5 \\
          & DANAA & 98.1 & \textbf{86.8} & \textbf{84.8} & \textbf{80.3} & \textbf{64.4} & \textbf{68.4} & \textbf{55.4} & \textbf{56.5} & \textbf{33.1} \\
    \hline
    \multirow{6}[2]{*}{Inc-v4} & MIM & 58.2 & \textbf{99.9} & 45   & 40.4 & 23.5 & 20.4 & 17.7 & 20.3 & 9.7 \\
          & NRDM & 78.0   & 96.4 & 62.8 & 62.3 & 26.1 & 25   & 17.3 & 16.6 & 6.8 \\
          & FDA  & 84.6 & 99.6 & 71.8 & 68.8 & 28.2 & 26.1 & 17.4 & 17.1 & 7.0 \\
          & FIA  & 74.6 & 91.0   & 69.6 & 65.7 & 43.5 & 47.3 & 39.3 & 39.9 & 23.5 \\
          & NAA  & 83.3 & 95.8 & 77.9 & 73.3 & 49.5 & 53.2 & 48.0   & 46.5 & 31.4 \\
          & DANAA & \textbf{86.8} & 97.2 & \textbf{82.4} & \textbf{76.9} & \textbf{54.9} & \textbf{61} & \textbf{53.8} & \textbf{53.5} & \textbf{35} \\
    \hline
    \multirow{6}[2]{*}{IncRes-v2} & MIM & 60 & 51.9 & \textbf{99.2} & 42.2 & 25.9 & 30.5 & 21.7 & 23.3 & 12.3 \\
         & NRDM & 72.8 & 67.9 & 77.9 & 59.7 & 35.7 & 30.8 & 16.4 & 17.1 & 7.3 \\
         & FDA & 69.0 & 68.0 & 78.2 & 56.2 & 34.5 & 29.7 & 16.2 & 15.4 & 7.7 \\
         & FIA & 71.0 & 68.2 & 78.8 & 63.9 & 53.8 & 56.4 & 47.4 & 45.8 & 37.6 \\
         & NAA & 79.5 & 76.4 & 89.3 & 71.1 & 60.3 & 64.8 & 56.9 & 55.0 & 47.3 \\
         & DANAA & \textbf{82.7} & \textbf{80.4} & 91.5 & \textbf{77.7} & \textbf{66.3} & \textbf{72.2} & \textbf{64.7} & \textbf{60.8} & \textbf{56} \\
    \hline
    \multirow{6}[2]{*}{Res152-v2} & MIM & 52.9 & 47.3 & 44.9 & \textbf{99.4} & 26.6 & 25.1 & 24.3 & 24.4 & 13.3 \\
        & NRDM & 72.7 & 68.8 & 59.5 & 89.9 & 39.1 & 31.0 & 20.3 & 18.1 & 9.3 \\ 
        & FDA & 15.7 & 9.2 & 8.3 & 26.2 & 13.1 & 6.8 & 9.3 & 9.7 & 4.0 \\ 
        & FIA & 80.7 & 78.2 & 77.5 & 98.0 & 58.5 & 58.2 & 53.0 & 48.4 & 34.4 \\ 
        & NAA & 84.7 & 83.5 & 82.3 & 97.6 & 61.8 & 67.0 & 59.1 & 58.1 & 46.1 \\ 
        & DANAA & \textbf{86.4} & \textbf{86.8} & \textbf{85.9} & 98.8 & \textbf{68.1} & \textbf{71.7} & \textbf{65.1} & \textbf{62.0} & \textbf{48.4} \\ 
    \hline
    \multirow{6}[2]{*}{Inc-v3} & MIM-PD & \textbf{99.7} & 72.8 & 66.9 & 54.1 & 31.7 & 29.1 & 20.2 & 21.7 & 9.7 \\
        & NRDM-PD & 86.3 & 68.6 & 64.3 & 58.0 & 31.1 & 22.6 & 10.6 & 13.8 & 5.9 \\
        & FDA-PD & 74.7 & 49.3 & 46.5 & 40.9 & 23.7 & 15.4 & 10.5 & 13.1 & 6.2 \\
        & FIA-PD & 96.9 & 83.5 & 82.7 & 79.8 & 61.4 & 62.1 & 47.0 & 48.2 & 27.5 \\
        & NAA-PD & 97.2 & 87.0 & 85.6 & 81.1 & 64.9 & 65.8 & 53.4 & 51.6 & 33.6 \\
        & DANAA-PD & 97.9 & \textbf{89.4} & \textbf{89.4} & \textbf{84.8} & \textbf{70.6} & \textbf{72.3} & \textbf{61.7} & \textbf{60.9} & \textbf{40.1} \\
    \hline
    \multirow{6}[2]{*}{Inc-v4} & MIM-PD & 81.3 & \textbf{99.4} & 71.0 & 59.7 & 31.6 & 28.0 & 22.9 & 23.3 & 12.7 \\
        & NRDM-PD & 90.3 & 97.0 & 79.5 & 76.8 & 34.1 & 34.4 & 21.1 & 19.7 & 8.6 \\
        & FDA-PD & 93.2 & 99.2 & 86.4 & 82.4 & 36.7 & 37.4 & 20.3 & 21.1 & 10.0 \\
        & FIA-PD & 84.0 & 92.4 & 81.2 & 77.1 & 55.2 & 58.6 & 48.9 & 47.5 & 29.3 \\
        & NAA-PD & \textbf{90.5} & 96.9 & 87.6 & 83.9 & 58.4 & 64.3 & 54.0 & 53.4 & 34.6 \\
        & DANAA-PD & 90.4 & 96.5 & \textbf{87.9} & \textbf{84.9} & \textbf{63.9} & \textbf{71.0} & \textbf{61.9} & \textbf{60.1} & \textbf{42.7} \\
    \hline
    \multirow{6}[2]{*}{IncRes-v2} & MIM-PD & 80.7 & 76.5 & \textbf{98.0} & 65.8 & 36.9 & 42.7 & 29.4 & 28.6 & 17.1 \\
        & NRDM-PD & 76.4 & 74.1 & 78.7 & 64.1 & 40.7 & 32.4 & 17.5 & 18.8 & 6.7 \\
        & FDA-PD & 78.1 & 76.2 & 80.7 & 66.5 & 41.3 & 35.6 & 18.4 & 17.0 & 7.6 \\
        & FIA-PD & 76.5 & 73.4 & 81.7 & 71.1 & 60.0 & 62.5 & 50.3 & 47.0 & 36.4 \\
        & NAA-PD & 81.4 & 78.2 & 89.9 & 76.4 & 65.2 & 67.7 & 59.9 & 57.1 & 46.0 \\
        & DANAA-PD & \textbf{83.7} & \textbf{80.4} & 89.8 & \textbf{80.6} & \textbf{70.3} & \textbf{73} & \textbf{65.8} & \textbf{63.1} & \textbf{55.8} \\
    \hline
    \multirow{6}[2]{*}{Res152-v2} & MIM-PD & 81.5 & 77.5 & 76.2 & \textbf{99.4} & 41.5 & 44.5 & 34.8 & 33.6 & 18.4 \\
        & NRDM-PD & 84.1 & 82.1 & 73.1 & 90.1 & 51.6 & 43.5 & 28.3 & 22.5 & 11.2 \\
        & FDA-PD & 22.1 & 12.7 & 11.4 & 23.4 & 19.6 & 10.4 & 9.9 & 11.7 & 5.4  \\
        & FIA-PD & 88.6 & 86.1 & 87.0 & 98.3 & 70.9 & 71.0 & 63.6 & 58.6 & 43.4 \\
        & NAA-PD & 90.2 & 88.5 & 89.0 & 98.0 & 73.5 & 76.1 & 70.3 & 66.3 & 52.2 \\
        & DANAA-PD & \textbf{92.0} & \textbf{91.7} & \textbf{91.8} & 98.7 & \textbf{79.3} & \textbf{82.1} & \textbf{76.1} & \textbf{73.4} & \textbf{60.8} \\
    \hline
  \bottomrule
  \end{tabular}
\end{table}

\subsection{Ablation Study}
In this section, we investigate the impact of the learning rate and Gaussian noise deviations on the performance of the proposed method.


\subsubsection{The Impact of Learning Rates.}
\begin{figure}[htbp]
  \centering
  \includegraphics[width=0.95\textwidth]{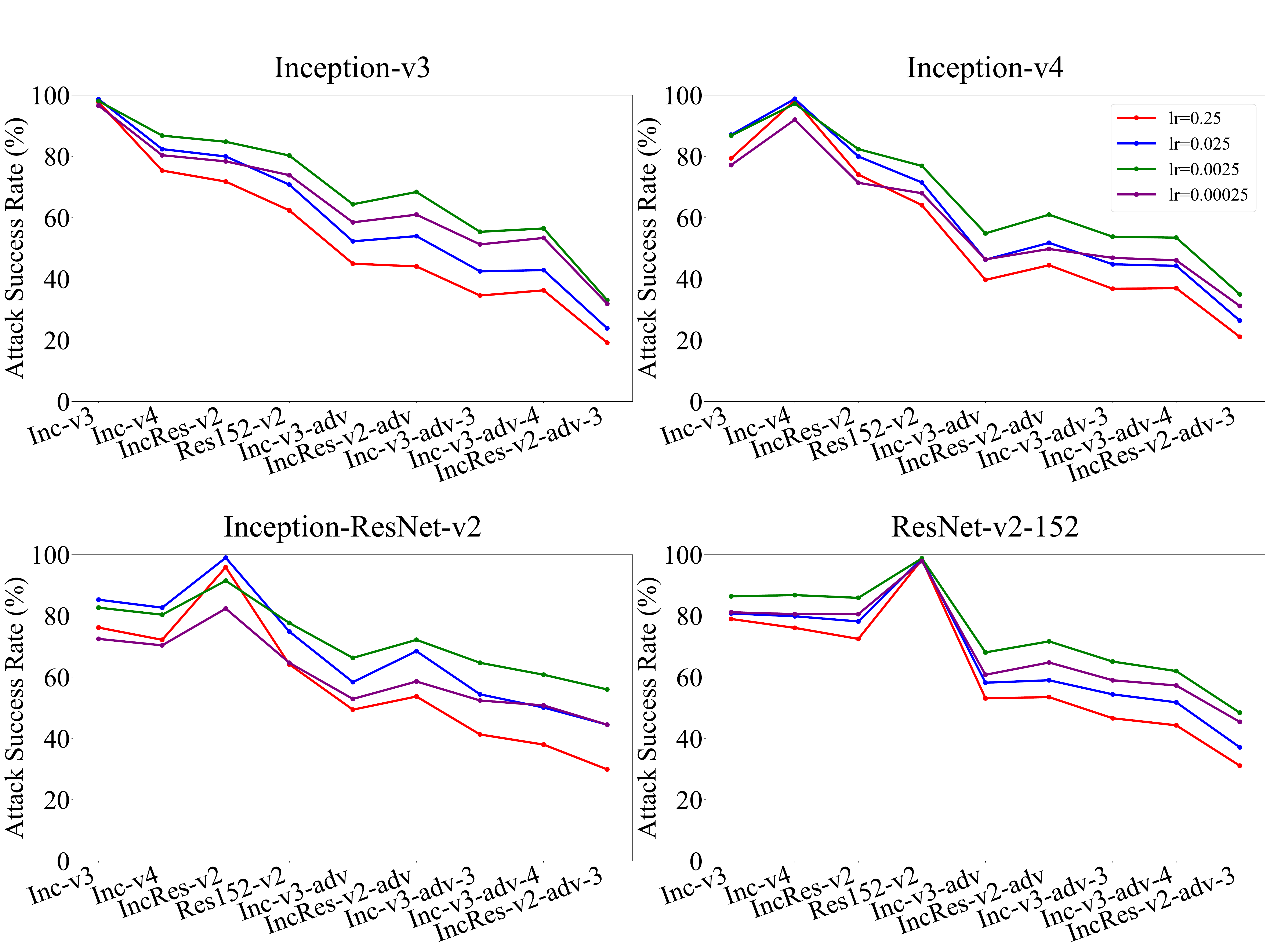}
  \caption{DANAA attack success rate performance at different learning rates}
  \label{fig:lr}
\end{figure}

\begin{figure}[htbp]
  \centering
  \includegraphics[width=0.95\textwidth]{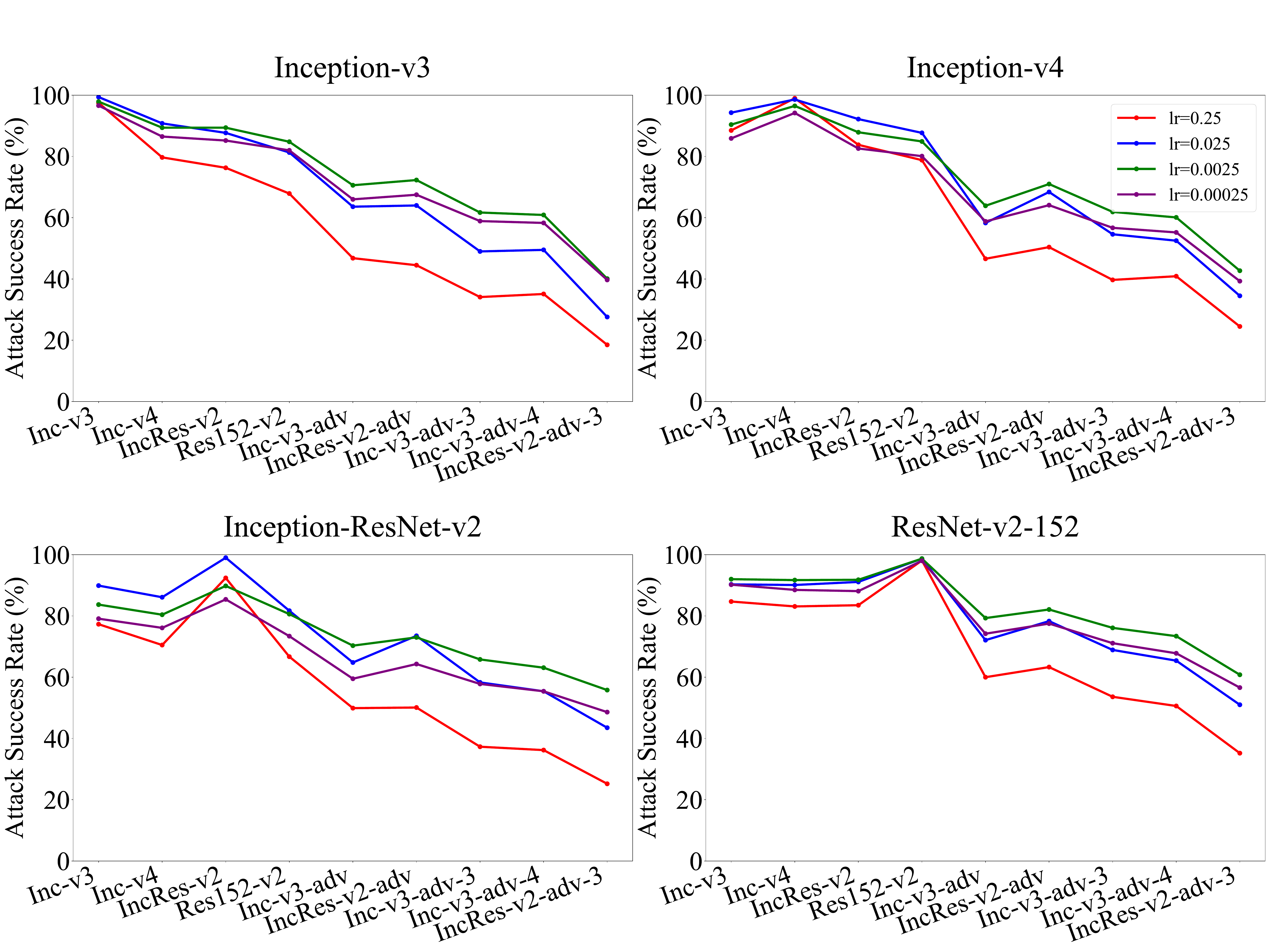}
  \caption{DANAA-PD attack success rate performance at different learning rates}
  \label{fig:lr-PD}
\end{figure}

Experiments are conducted using different scales of learning rates, which are 0.25, 0.025, 0.0025 and 0.00025. In Fig.~\ref{fig:lr}, the DANAA method exhibits the highest attack success rate for nearly all models when the selected learning rate was 0.0025. In Fig.~\ref{fig:lr-PD}, the highest attack success rates are achieved on most models for DANAA-PD method. 

Notably, when using Inception-ResNet-v2 as the source model, although at a learning rate of 0.0025 DANAA-PD ranked second best in attack success rate on the models without adversarial training, its effectiveness on the model with adversarial training is still much higher than those at other learning rates.

\subsubsection{The Impact of Gaussian Noise Deviation (Scale).}
\begin{figure}[htbp]
  \centering
  \includegraphics[width=0.95\textwidth]{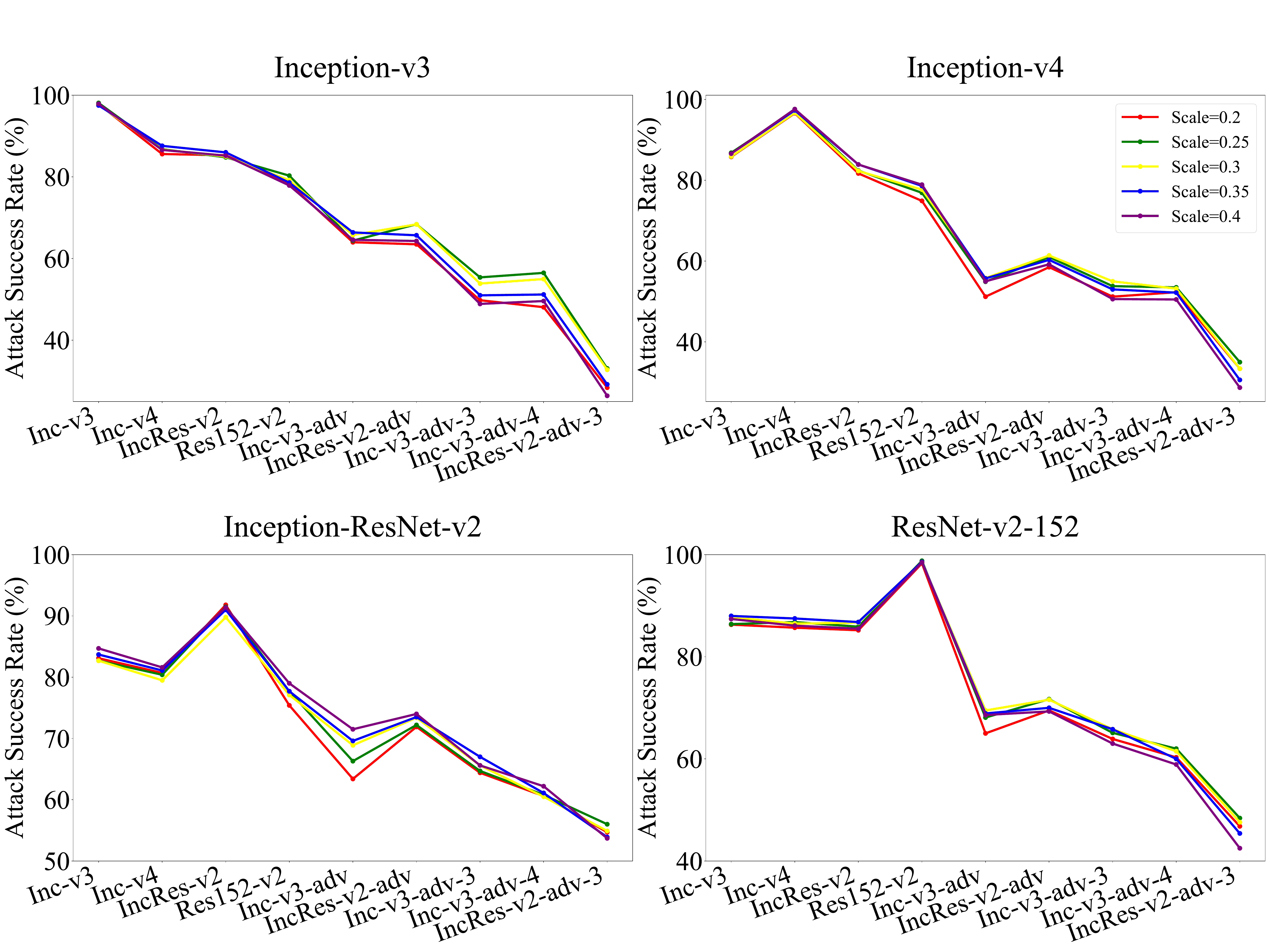}
  \caption{DANAA attack success rate performance at different noise deviation}
  \label{fig:scale}

\end{figure}

\begin{figure}[htbp]
  \centering
  \includegraphics[width=0.95\textwidth]{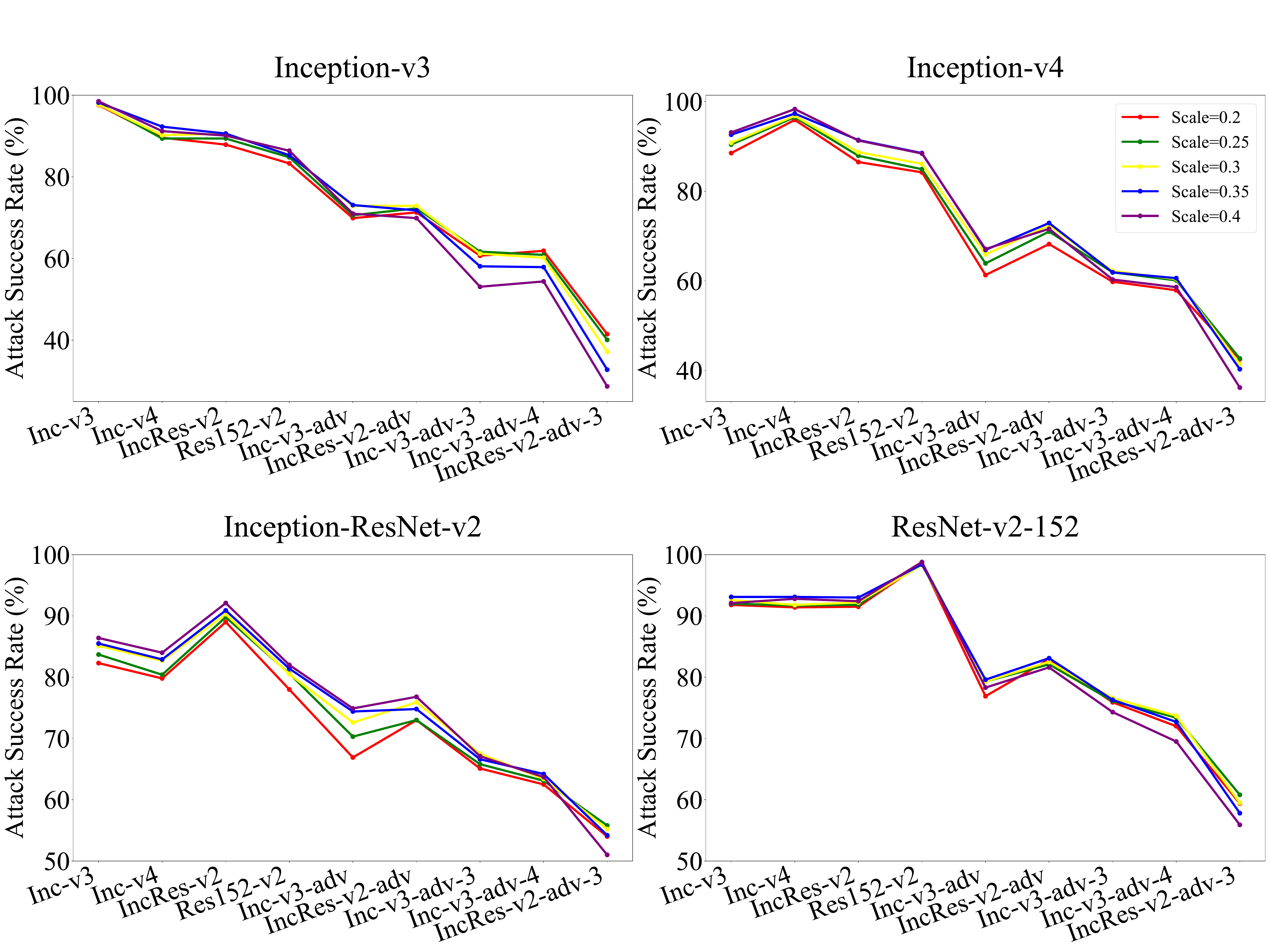}
  \caption{DANAA-PD attack success rate performance at different noise deviation}
  \label{fig:scale-PD}
\end{figure}

To verify the effect of adding Gaussian noise to the gradient update on model transferability in this paper, we selected different noise deviations for testing in this subsection. As shown in Fig.~\ref{fig:scale} and Fig.~\ref{fig:scale-PD}, five different scales of the Gaussian noise deviation ranging from 0.2 to 0.4 are used in this experiment. In general, higher value of scale tends to have more superior results for the normal training model while sacrificing performance for the more robust one. Conversely, a lower value of scale results in less improvements for the normal trained model but better performance for the adversarial trained model. Accordingly, the scale value of 0.25 is selected for the optimal performance in this paper.

\section{Conclusion}
In this paper, we propose a double adversarial neuron attribution attack method (DANAA) to achieve enhanced transfer-based adversarial attack results. Compared with other literature methods, our method obtains a better transferability for the adversarial samples. To derive more accurate importance estimates for the middle layer neurons, we firstly employ a non-linear path to the perturbation update process. Considering the calculation of gradient on the non-linear path, for all examined models, the performance of DANAA algorithm has substantially improved by up to 9.0\% in comparison with the second best method with adversarial trained models, and has an average overall improvement by 7.1\%. With the information transformation methods of DIM and PIM, our DANAA-PD algorithm also has a maximum enhancement of 9.8\% and an average overall improvement of 7.3\% compared to NAA-PD algorithm. Extensive experiments have demonstrated that the attribution model proposed in this paper achieves the state-of-the-art performance, with greater transferability and generalisation capabilities.

\bibliographystyle{splncs04}
\bibliography{mybibliography}
%




\end{document}